\documentclass[10pt,twocolumn,letterpaper]{article}

\usepackage{cvpr}

\usepackage[pagebackref,breaklinks,colorlinks,allcolors=cvprblue]{hyperref}
\usepackage{booktabs}
\usepackage{multirow}
\usepackage{hyperref}
\usepackage{url}
\usepackage{booktabs}
\usepackage{graphicx}
\usepackage[table]{xcolor}
\usepackage[dvipsnames]{xcolor}
\usepackage{enumitem}
\usepackage{subcaption}
\usepackage{makecell}
\usepackage{pifont}
\usepackage{multirow,makecell}
\usepackage[dvipsnames]{xcolor}
\usepackage{pifont}
\usepackage{amssymb}

\newcommand{\rengen}{RenGen}
\newcommand{\boldheader}[1]{\vspace{2.3pt}\noindent\textbf{#1.}}

\newcommand{\cmark}{\textcolor{ForestGreen}{\ding{51}}}
\newcommand{\xmark}{\textcolor{BrickRed}{\ding{55}}}
\newcommand{\gr}[1]{\textcolor{gray}{#1}}

\definecolor{cvprblue}{rgb}{0.21,0.49,0.74}
\definecolor{RowBlue}{RGB}{200, 230, 242}

\def\confName{CVPR}
\def\confYear{2027}

\title{\rengen{}: Generating Images with as Many Tokens as the Scene Needs}

\author{
    Savya Khosla \quad Jay Mahajan \quad Akin Tuna Sakalli \quad Derek Hoiem \quad David Forsyth \quad Alex Schwing \\
    \small University of Illinois Urbana-Champaign
}

\begin{document}
\maketitle
\begin{abstract}
We present \rengen{}, a latent diffusion model that generates simple scenes from a few tokens and busy scenes from many, saving time and compute on simple scenes without losing realism. Most latent generators represent every image as a fixed grid of patch tokens, so a simple image costs as much to generate as a complex one. Prior methods that vary the token count through content-adaptive tokenization lose image quality as the count shrinks. \rengen{} instead trades scene complexity, rather than image quality, for fewer tokens. It does so by generating in the latent space of a pretrained region encoder, where each token represents one region of the image, and a decoder renders these region tokens back into pixels. Unlike patch tokens, region tokens cover areas of varying shape and position, so each token must capture both what its region contains and where it lies on the image canvas. We encode this spatial extent by adding positional embeddings to the encoder features before pooling, and we generate it by anchoring each token to a single randomly sampled point, from which the generator infers the region's shape, size, and content. At sampling time, the number of anchor points sets the trade-off between scene complexity and compute. On ImageNet $256{\times}256$, \rengen{} reaches a gFID of 1.34 with 88 tokens per image on average, the best among methods that vary their token count. It approaches the quality of RAEv2, a state-of-the-art patch-based model, at $2.5{\times}$ lower latency, advancing the Pareto frontier of generation quality versus compute. \rengen{}'s gFID stays flat across scene complexity while latency varies by $2.4{\times}$ between simplest and busiest scenes, and its autoencoder reconstructs nearly as well as RAEv2 (rFID 0.32 vs.\ 0.30) with $2.9{\times}$ fewer tokens.
\end{abstract}
    
\section{Introduction}
\label{sec:intro}

\begin{figure}[t]
    \centering
    \includegraphics[width=\linewidth]{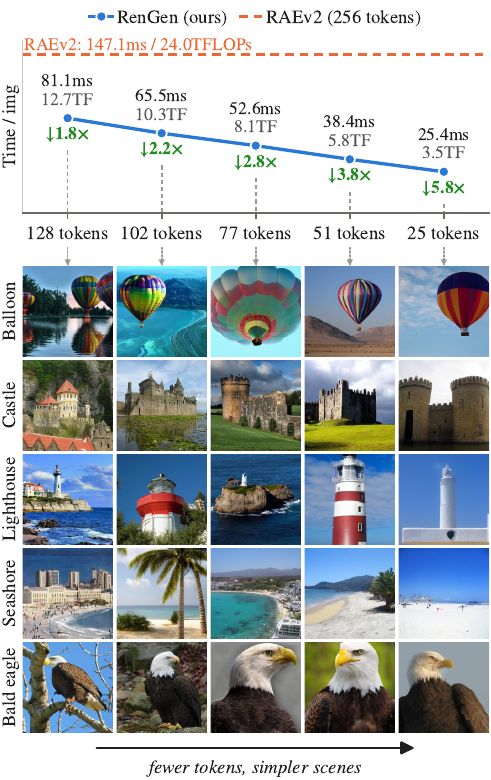}
    \vspace{-1.8em}
    \caption{{\bf \rengen{}'s generation cost follows image content.} {\em Top:} Wall-clock time and TFLOPs per image for end-to-end generation at different token counts for \rengen{}. Green values give the speed-up over RAEv2, which always uses a fixed grid of 256 patch tokens (dashed line). {\em Bottom:} $256{\times}256$ samples from a single \rengen{} model. Each row uses the same class label, and each column uses the token count of the point above it, so fewer tokens yield simpler scenes at lower cost.}
    \label{fig:teaser}
    \vspace{-0.6em}
\end{figure}

\begin{figure*}[t]
    \centering
    \includegraphics[width=\linewidth]{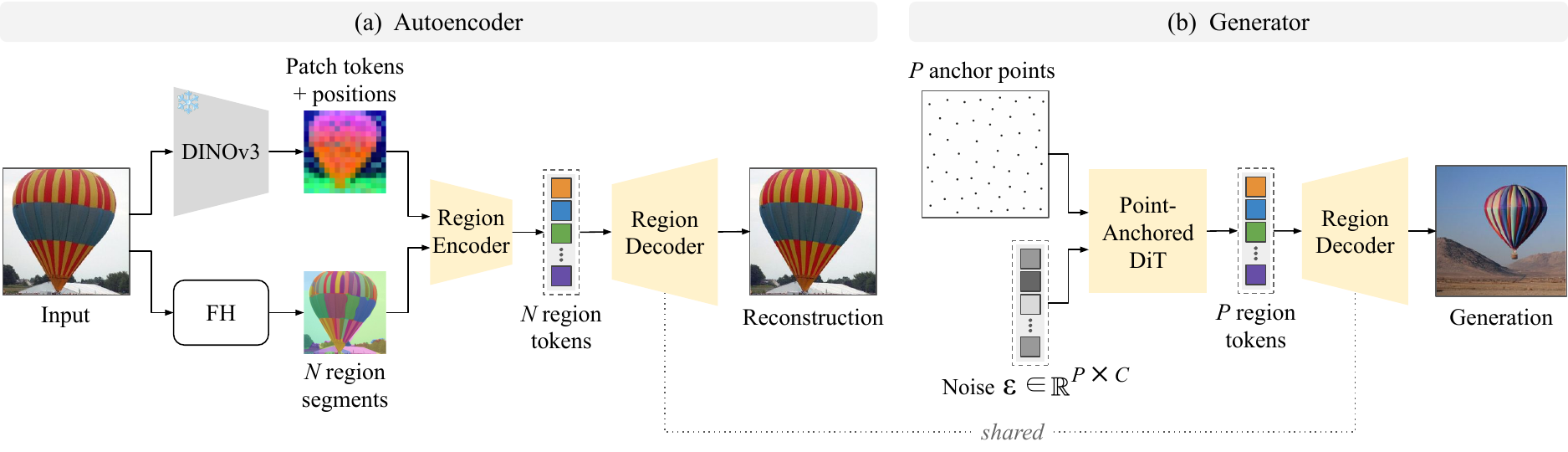}
    \vspace{-1.8em}
    \caption{{\bf \rengen{} overview.} (a) The region encoder partitions an image into $N$ regions with Felzenszwalb–Huttenlocher (FH) segmentation and pools frozen DINOv3 patch features within each region into region tokens, which the decoder renders back into an image. (b) For generation, $P$ random anchor points fix the region positions. A DiT denoises one region token per anchor point, using the anchor coordinates as 2D RoPE positions, and the shared decoder renders the generated tokens into an image.}
    \label{fig:architecture}
    \vspace{-0.6em}
\end{figure*}

We show how to generate realistic images from a small number of tokens, thereby reducing generation time and compute (Fig.~\ref{fig:teaser}). We do so by designing a generative model whose latents are region tokens from a pretrained image encoder, each representing one region of the image~\cite{RegionRevisited,REN,TREN}. The number of tokens then controls scene complexity rather than image quality: fewer tokens yield simpler scenes and more tokens yield busier ones, both equally realistic. The key difficulty in generating with region tokens is that each token must specify which pixels it covers in addition to what it contains. We show that anchoring each token to a single randomly sampled point suffices for the generator to infer the region’s full extent (shape and size) and content.

Most current generators tokenize every image into a uniform grid of patches~\cite{LDM,DiT,RAE,JiT}, requiring the same generation time and compute for simple and complex images at a given resolution (dashed line in Fig.~\ref{fig:teaser}). Generators whose compute instead scales with scene complexity would let a single model serve devices with different compute budgets and latency requirements. Prior work pursues this with content-adaptive tokenizers~\cite{FlexTok,OneDPiece,CAT} and pixel-space generators that use a variable number of tokens~\cite{kara2026budgetpix,RTI}. In both, however, fewer tokens come at the cost of quality. Adaptive tokenizers lag behind leading patch-based autoencoders in reconstruction (Tab.~\ref{tab:comparison}), and variable-token generators lose image quality as the token count shrinks (Fig.~\ref{fig:complexity}).

We introduce \rengen{}, a latent generative model built on region tokens. \rengen{} derives these tokens from the patch features of a pretrained image encoder such as DINOv3~\cite{DINOv3}, which RAE~\cite{RAE} identifies as effective diffusion latents. Following region encoders~\cite{RegionRevisited,REN}, it pools these features within image segments, so that the token count follows image content. A decoder renders the resulting region tokens into pixels, and a diffusion transformer (DiT)~\cite{DiT} generates them from noise (Fig.~\ref{fig:architecture}). Since a region token covers an arbitrary set of pixels, both  decoder and  generator face a design question that patch-based models avoid.

\noindent\textbf{How can the decoder recover a region’s spatial extent from a single token?} Region tokens from prior region encoders~\cite{RegionRevisited,REN,TREN} do not retain enough positional information to decode a region’s spatial extent (Sec.~\ref{sec:encoding-layout}). \rengen{} addresses this by adding random Fourier positional embeddings to the patch features before pooling, so that each region token records where its region lies along with what it contains. The decoder then reconstructs the image from patch-grid queries that carry Fourier features of the same random directions and cross-attend to the region tokens, so that each output location learns to gather detail from the regions that cover it, regardless of their number or layout.

\noindent\textbf{How can the DiT generate region tokens when nothing fixes where each region lies?} Unlike patch-based generators, where each token's position is fixed in advance to a $16{\times}16$ grid cell and the model only needs to generate its content, a region-token generator lacks any implicit spatial prior. We find that generating region layouts jointly with content degrades quality, while generating them in a separate stage erodes the efficiency gained from fewer tokens (Sec.~\ref{sec:generating-layout}). Instead, we anchor each token to a single point sampled uniformly on the image canvas, which serves as its 2D RoPE position. From this anchor point our DiT generates the region’s shape, size, and content. The number of anchors $P$ thus becomes a sampling-time control over the trade-off between scene complexity and compute.

In summary, our contributions are:
\begin{itemize}
    \item We introduce \rengen{}, a latent diffusion model over region tokens that trades scene complexity, not image quality, for compute. Its token budget is chosen at sampling time, and on ImageNet $256{\times}256$ a single model reaches a gFID of 1.34 with 88 tokens on average, the best among methods that vary their token count, at $2.5{\times}$ lower latency than an identically configured patch-based model.
    \item We make region tokens decodable into images. Tokens formed by simply pooling pretrained image encoder features, as in prior region encoders~\cite{RegionRevisited,REN,TREN}, carry too little position information for the decoder to recover which pixels each one covers. Studying several designs, we find a simple fix that future region encoders can readily adopt, which is to add positional embeddings to the patch features before pooling them into region tokens. The resulting model reconstructs nearly as well as RAEv2 (rFID 0.32 vs.\ 0.30) with $2.9{\times}$ fewer tokens.
    \item We show that placing each region token at a random point on the image is enough for the generator to produce realistic regions around it. Unlike patch tokens, region tokens have no fixed grid that tells the generator where each one belongs, and generating their positions alongside their content is either slow or degrades quality. Anchoring region token generation to random points avoids both problems at no extra cost.
\end{itemize}    
\section{Related Work}
\label{sec:related}

\boldheader{Region encoders}
Recent works show that region tokens pooled from pretrained vision foundation models are a strong alternative to patch tokens~\cite{RegionRevisited,REN,TREN}. These representations use far fewer tokens and perform well across tasks, including semantic segmentation~\cite{OVSeg,ODISE,TextRegion,REN}, one-shot object localization and retrieval~\cite{Matcher,RELOCATE}, vision-language modeling~\cite{GLAMM,Osprey,RegionGPT,RVLM}, and video understanding~\cite{RELOCATE,VideoRefer}. While effective for understanding tasks, we find that region tokens from prior methods are insufficient for reconstruction, because pooled tokens carry little information about where a region lies and what shape it has. We make region tokens decodable by adding positional embeddings to patch features before pooling. Object-centric models also decode images from region-like latents~\cite{OCSD,SlotDiffusion}, but they use a fixed number of slots per image, whereas our token count follows the number of segments in each image.

\boldheader{Representation-based generation}
Recently, there has been a push towards training diffusion transformers with representation autoencoders (e.g., DINO, SigLIP, MAE)~\cite{RAE, RAEv2, chen2025maetok} instead of standard VAEs~\cite{kingma2022autoencodingvariationalbayes} used in modern text-to-image and text-to-video models~\cite{podell2023sdxlimprovinglatentdiffusion, flux2024, yang2025cogvideoxtexttovideodiffusionmodels, hacohen2024ltxvideo}. These works show the promise of improved generation as these autoencoders are believed to have a better semantic understanding of images. Our work follows a similar motivation as we use a representation encoder, but aim to reduce the required token count by focusing on the regions of the images themselves rather than the patches.

\boldheader{Efficient token generation}
Several works explored efficient ways to generate an image with fewer tokens. RTI \cite{RTI} employs a Hilbert curve to detect which features change the most and uses LoRA fine-tuning to handle variable region token budgets. Token Merging (ToMe) \cite{bolya2023tomesd} groups tokens by feature similarity within a forward pass. BudgetPix \cite{kara2026budgetpix} employs an adaptive encoder and decoder with multiple resolution paths and fine-tunes an existing DiT. All of these works either adapt an existing pixel-based or latent-based flow model to complete this task. Our work differs as we directly model a region space that is semantically aligned and flexible to handle a variable number of tokens.
    
\section{\rengen{}}
\label{sec:method}

\begin{figure}[t]
    \centering
    \includegraphics[width=\linewidth]{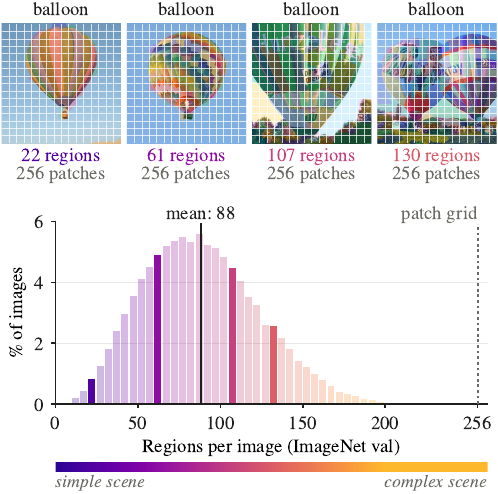}
    \vspace{-1.8em}
    \caption{{\bf Image complexity varies, but patch tokenizers spend the same budget on every image.} {\em Top:} ImageNet images of increasing complexity, overlaid with the $16{\times}16$ patch grid of patch-based latent diffusion models and with their region partitions. The simplest of the four scenes has only 22 regions, while the busiest has 130. {\em Bottom:} Regions per image across ImageNet validation set. The mean of 88 is $2.9{\times}$ fewer than the 256 tokens of a patch grid (dashed line).}
    \label{fig:regions-histogram}
\end{figure}

Images vary widely in complexity, yet patch-based latent generative models represent every image with the same number of tokens (Fig.~\ref{fig:regions-histogram}). \rengen{} instead uses a variable-size latent representation that adapts to image content. \rengen{} partitions each image into $N$ regions and represents each with a region token obtained by pooling frozen DINOv3 features within that region (Sec.~\ref{sec:encoder}). A decoder maps these tokens back to pixels through a grid of patch queries that cross-attend to the region tokens (Sec.~\ref{sec:decoder}). For generation, we place $P$ anchor points on the image canvas and use a DiT to generate one region token per anchor, which the decoder renders into an image (Sec.~\ref{sec:dit}). The number of anchors $P$ sets both the complexity of the generated image and the generator's compute. Fig.~\ref{fig:architecture} shows an overview of \rengen{}.

\subsection{Region Encoder}
\label{sec:encoder}

\boldheader{Content-adaptive partitioning} 
We partition each image into regions using the Felzenszwalb--Huttenlocher (FH) algorithm~\cite{Felzenszwalb}. FH greedily merges pixels into segments whose internal color variation is small relative to the differences across their boundaries. We run FH with scale $k{=}100$, Gaussian smoothing $\sigma{=}0.8$, and a minimum segment size of 32 pixels. The resulting segments are then aligned to the backbone's patch grid. Each patch is owned by the smallest segment covering at least 25\% of its area, or otherwise by the segment covering the largest patch area. Any segment that owns no patches is merged into the owners of the patches it overlaps. The resulting regions average 88 per image on the ImageNet validation set, with the 5th and 95th percentiles at 34 and 151, respectively. For each region, we store a coverage mask $m_i$ that records the fraction of each patch the region covers.

\boldheader{Pooling patch features into region tokens} 
We use a lightweight cross-attention module to pool frozen DINOv3~\cite{DINOv3} patch features within each region. Following RAEv2~\cite{RAEv2}, we use DINOv3 features from seven layers, sampled at every other layer starting from the final one. We add random Fourier positional embeddings to these features before pooling, so that each region token records both what its region contains and where it lies. The cross-attention query for each region starts as the mean of its patch features weighted by the coverage mask $m_i$ and is refined by $L_e{=}4$ cross-attention blocks. Attention logits are biased by $\log m_i$, which restricts each query to its region and down-weights partially covered patches. We ablate different ways of encoding position in the token in Sec.~\ref{sec:encoding-layout}.

\subsection{Region Decoder}
\label{sec:decoder}

\boldheader{Query grid} 
Unlike patch tokens, region tokens vary in number and have no fixed location in the output image, yet they must be decoded into an image of a specified resolution. To do this, the decoder places one query on each $16{\times}16$ output patch, with each query reading from the region tokens through cross-attention. A $G{\times}G$ query grid therefore produces a $16G{\times}16G$ image, making the output resolution independent of the number of region tokens. Each query is initialized with Fourier features of its patch position, averaged over the patch area to suppress frequencies finer than the patch grid.

\boldheader{Decoding by cross-attention} 
The decoder consists of $L_d$ transformer layers ($L_d{=}6$ for ViT-B and $14$ for ViT-XL). Each layer applies cross-attention from the queries to the region tokens, followed by self-attention among the queries, with an MLP after each attention operation. Following DETR~\cite{detr}, we add each query's Fourier features before every attention operation. These features share their random directions with the Fourier embeddings added to the patch features in the encoder, which helps cross-attention associate each query with the regions covering its position. Finally, a linear head maps each query to the RGB values of its corresponding output patch.

\subsection{Autoencoder Training}
\label{sec:ae-training}

We train the encoder's pooling module end-to-end with the decoder on the reconstruction objective. The DINOv3 backbone is kept frozen. Given an image $x$ and its reconstruction $\hat{x}$, we minimize
\begin{equation}
    \mathcal{L} = \lVert x - \hat{x} \rVert_1 + \lambda_{\text{p}}\, \mathcal{L}_{\text{LPIPS}}(x, \hat{x}) + \lambda_{\text{adv}}\, \mathcal{L}_{\text{adv}}(\hat{x}),
\label{eq:ae-loss}
\end{equation}
where $\mathcal{L}_{\text{LPIPS}}$ is the perceptual loss~\cite{LPIPS} and $\mathcal{L}_{\text{adv}}$ is the adversarial loss. The training recipe follows RAEv2~\cite{RAEv2}, with details in Appendix.

\begin{figure}[t]
    \centering
    \includegraphics[width=\linewidth]{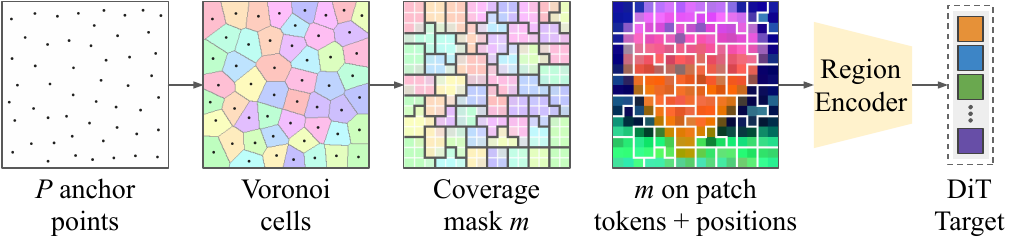}
    \vspace{-1.8em}
    \caption{{\bf Constructing DiT targets.} During training, we place $P$ anchor points on the image canvas, partition the canvas into their Voronoi cells, compute each cell's coverage mask on the patch grid, and encode the image with the region encoder to obtain one target region token per anchor.}
    \label{fig:dit-targets}
\end{figure}

\subsection{Point-Anchored DiT}
\label{sec:dit}

Unlike patch tokens, region tokens have no fixed number, positions, sizes, or shapes, so a generator must decide both the {\em spatial layout} (the position, shape, and size of each region) and the {\em content} (the semantics and appearance of each region). We explore several ways of generating layout and content (Sec.~\ref{sec:generating-layout}) and find that conditioning on real layouts requires training images at sampling time and leads to memorization, generating layouts jointly with content underperforms, and generating them with a separate DiT adds a second sequential stage that erodes the efficiency gained from generating fewer tokens. We instead fix only the region positions before generation, as described next.

\boldheader{Placing anchor points} 
We fix a coarse layout by placing $P$ image-independent anchor points on the image canvas, sampled uniformly and spread evenly with three iterations of Lloyd’s algorithm~\cite{Lloyd}. During training, these anchors define the DiT targets (Fig.~\ref{fig:dit-targets}). Concretely, we (1) set $P$ to the image’s FH region count randomly varied by $\pm20\%$ and clipped to $[16, 196]$; (2) place $P$ anchor points on the image canvas and partition it into their Voronoi cells; (3) compute a coverage mask for each cell on the backbone’s patch grid; and (4) encode the image with the region encoder, yielding one region token per anchor as the DiT’s target. During sampling, we set $P$ to the FH region count of a random training image of the target class (88 on average), or choose it manually to trade scene complexity for compute. 

\boldheader{Denoising region tokens at anchor points} 
We use a DiT with a DDT head following RAE~\cite{RAE}. Its input is the set of $P$ noisy tokens, with no patch grid. Each token's only spatial input is its anchor point, which we use as 2D RoPE~\cite{RoPE} in every attention layer. The timestep, the class label, and the token count $P$ enter as additional in-context tokens. During sampling, we integrate the ODE with $T{=}50$ Euler steps under the timestep shift described below, using internal guidance~\cite{IG} from an auxiliary head on an intermediate block. The region decoder (Sec.~\ref{sec:decoder}) renders the image from the denoised region tokens.

\boldheader{Training} 
We normalize the target tokens per channel and train with flow matching~\cite{FlowMatching} on the interpolant $x_t = (1-t)\,x + t\,\epsilon$, predicting $x$ and computing the loss on the implied velocity. To batch token sets of different sizes, we pad them to a common length, mask the padded tokens in attention, and average the loss over real tokens only. We drop the class label with probability 0.1 to enable classifier-free guidance~\cite{ho2021classifier}. We sample $t$ from a logit-normal distribution and apply the resolution-dependent shift $t \mapsto st/(1+(s-1)t)$ with $s{=}8$.    
\section{Experiments}
\label{sec:experiments}

\begin{figure}[t]
    \centering
    \includegraphics[width=\linewidth]{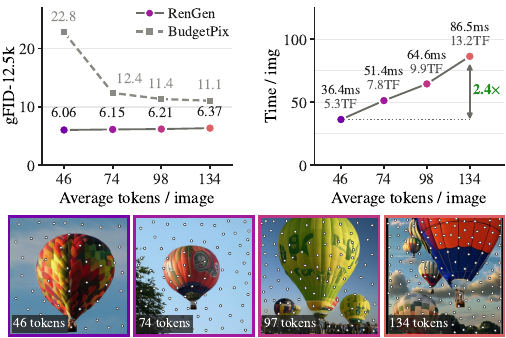}
    \vspace{-1.8em}
    \caption{{\bf \rengen{} adapts its compute to scene complexity without losing quality.} gFID stays flat across four complexity buckets, so in low-resource settings \rengen{} can generate simple, realistic scenes $2.4{\times}$ faster and with $2.5{\times}$ less compute than complex ones. Samples of the same class (bottom; anchor points in white) grow busier as the token budget increases.}
    \label{fig:complexity}
    \vspace{-0.8em}
\end{figure}

We first show that \rengen{} adapts its compute to scene complexity while maintaining image quality (Sec.~\ref{sec:complexity}). We then ablate how region tokens encode spatial layout (Sec.~\ref{sec:encoding-layout}) and how region layout and content are generated (Sec.~\ref{sec:generating-layout}). Next, we compare \rengen{} with representation autoencoders (RAEs) and fixed-length, variable-length, and content-adaptive tokenizers on reconstruction and generation (Sec.~\ref{sec:comparison}). Finally, we show that \rengen{} scales to higher resolutions more efficiently than fixed-grid models (Sec.~\ref{sec:scaling-resolution}). All experiments use ImageNet~\cite{ImageNet} at $256{\times}256$. Implementation details, evaluation metrics, and the timing protocol are provided in the Appendix.

\begin{figure}[t]
    \centering
    \includegraphics[width=\linewidth]{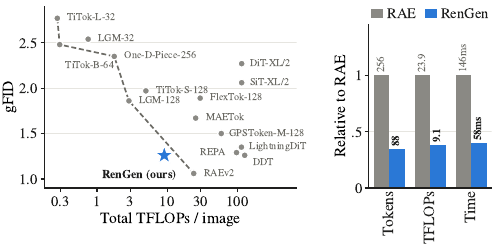}
    \vspace{-1.8em}
    \caption{{\bf \rengen{} sets a new quality--compute trade-off.} {\em Left:} gFID versus total TFLOPs per image on ImageNet $256{\times}256$. \rengen{} lies below the Pareto frontier of prior methods (dashed). {\em Right:} The source of this gain. At matched model size, \rengen{} generates $2.9{\times}$ fewer tokens than RAEv2, which translates directly into $2.6{\times}$ fewer TFLOPs and $2.5{\times}$ less wall-clock time per image.}
    \label{fig:pareto}
    \vspace{-0.8em}
\end{figure}

\subsection{Compute and quality across scene complexity}
\label{sec:complexity}
We show that \rengen{} allocates compute in proportion to scene complexity without sacrificing image quality. We use FH region count~\cite{Felzenszwalb} as a proxy for scene complexity because FH places a boundary wherever the color difference across it exceeds the variation within the adjacent segments, so a single object on a plain background yields few regions, while a cluttered scene yields many. 

We split the 50k ImageNet validation images into four equal-sized buckets by FH region count~\cite{Felzenszwalb}. For each bucket, \rengen{} generates 12.5k images, drawing class labels from the bucket's class distribution and $P$ from its region-count distribution. We compute gFID against the corresponding real-image bucket, ensuring that generated images are evaluated against scenes of similar complexity (Fig.~\ref{fig:complexity}). From the most to the least complex bucket, the average token count drops from 134 to 46, reducing TFLOPs by $2.5{\times}$ and wall-clock time per image by $2.4{\times}$. Despite this reduction, gFID remains nearly constant, ranging from 6.37 to 6.06. Thus, \rengen{} generates simpler scenes $2.4{\times}$ faster than complex ones at comparable image quality. In contrast, BudgetPix~\cite{kara2026budgetpix}, evaluated under the same protocol with matched token counts, sees its gFID degrade from 11.06 to 22.82 as its budget shrinks, indicating that it trades image quality, rather than scene complexity, for compute.

We also verify that the generated images reflect the complexity of their target buckets. The samples shown in Fig.~\ref{fig:complexity} illustrate this trend: scenes become progressively busier as the token count increases, while remaining realistic. Quantitatively, the mean FH region count of generated images is within 8\% of that of real images in every bucket, and within 3\% in all but the busiest bucket. It also correlates strongly with $P$ (Spearman $\rho = 0.94$), confirming that the number of anchor points largely determines the complexity of the generated scene.

\subsection{How to encode spatial extent in region tokens?}
\label{sec:encoding-layout}

A region token must carry both {\em what} a region contains and {\em where} it is, so the decoder can reconstruct the image by placing its content at the correct spatial location. We compare region tokens from a leading region encoder, REN~\cite{REN}, with three ways of constructing them ourselves (Tab.~\ref{tab:ae-ablation}).

\boldheader{REN tokens} Training a decoder on frozen REN tokens yields poor reconstructions, which we attribute to two properties of these tokens: (1) REN adds no positional encoding to the patch features it pools, so each token captures its region's extent only weakly via the image encoder's position embedding; and (2) its regions often overlap, so pixels are ambiguously assigned across tokens. Fine-tuning REN's cross-attention jointly with the decoder improves reconstruction only slightly, but the regions lose their spatial coherence and the number of tokens grows sharply during training.

\boldheader{Pooling patch features within FH segments} To fix the region count and obtain non-overlapping tokens, we replace REN's learned regions with Felzenszwalb--Huttenlocher (FH) segments~\cite{Felzenszwalb}, following RVLM~\cite{RVLM}, since FH is fast and covers every pixel with exactly one region. We pool DINOv3 features within each segment using the cross-attention module of Sec.~\ref{sec:encoder}, without positional embeddings. This resolves REN's overlap but not its weak position signal, because DINOv3 encodes position only through rotary embeddings in attention~\cite{RoPE,DINOv3}, leaving its patch features with little absolute position information. Reconstruction quality therefore still lags behind RAEv2 (Tab.~\ref{tab:ae-ablation}).

\boldheader{Explicit mask encoding} To strengthen the position signal, we add a learned linear projection of each region's mask to its pooled token. This closes most of the reconstruction gap to RAEv2, but generation suffers (Tab.~\ref{tab:ae-ablation}). The decoder learns to read each region's exact extent from the mask code, which simplifies reconstruction but shifts the burden to the DiT, since it must now generate these mask codes in addition to each region's content. We find that the DiT struggles to generate mask codes precise enough for the decoder to place regions correctly, so small errors in the codes become visible errors in the generated layout (Sec.~\ref{sec:generating-layout}).

\boldheader{Pooling patch features with embedded position (ours)} Instead of a separate mask code, we add random Fourier positional embeddings to the patch features before pooling, so that each region token encodes where its region lies together with what it contains. The decoder queries carry Fourier features of the same random directions, so cross-attention can match each query to the regions covering its position rather than reading their extent from an explicit code. This matches the reconstruction quality of explicit mask encoding without extra parameters. Because position is now part of the token content, the DiT no longer has to generate a separate, precise layout code, and it generates substantially better images.

\begin{figure*}[t]
    \centering
    \includegraphics[width=0.87\linewidth]{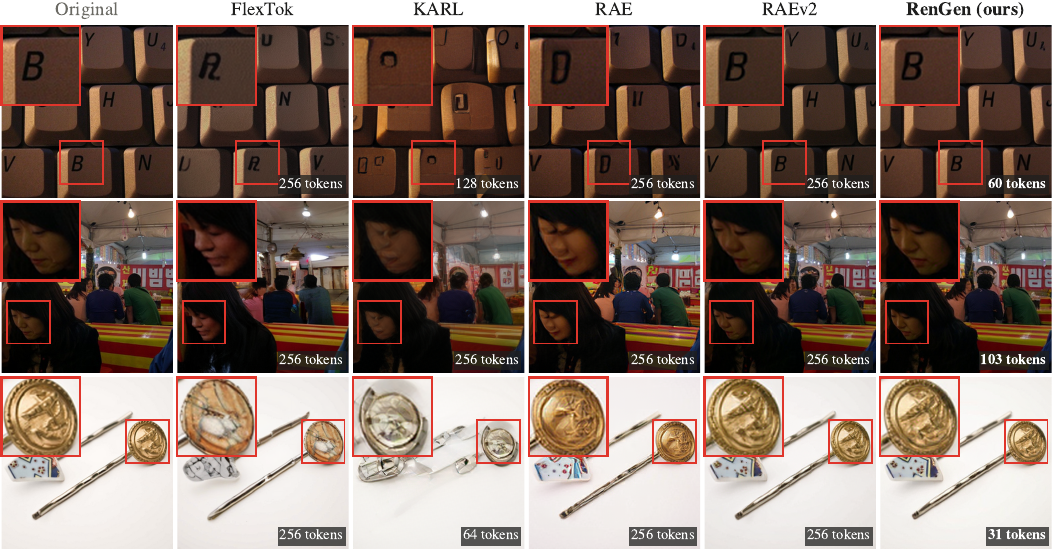}
    \caption{{\bf \rengen{} reconstructs fine detail as faithfully as RAEv2 with far fewer tokens.} Each tile shows its token count; \rengen{}'s follows image complexity, from 31 tokens for the hairpin on a plain background to 103 for the crowded scene.}
    \label{fig:reconstruction}
    \vspace{-0.6em}
\end{figure*}

\begin{table}[t]
    \centering
    \small
    \setlength{\tabcolsep}{4pt}
    \resizebox{\linewidth}{!}{
    \begin{tabular}{llccc}
    \toprule
    \textbf{Region tokens} &
    \textbf{Regions} &
    \textbf{Tokens} &
    \textbf{rFID}$\downarrow$ &
    \textbf{gFID}$\downarrow$ \\
    \midrule
    \gr{RAEv2 (for reference)} & \gr{n/a} & \gr{256} & \gr{0.30} & \gr{1.06} \\
    \midrule
    REN, frozen~\cite{REN}                 & Learned & 77                & 0.86 & -- \\
    REN, fine-tuned~\cite{REN}             & Learned & $77{\to}185^\ast$ & 0.47 & -- \\
    \midrule
    Pooling features without position      & FH      & 88                & 0.51 & -- \\
    Explicit mask encoding                 & FH      & 88                & \textbf{0.32} & 3.10 \\
    \rowcolor{RowBlue}
    Pooling patch features with position   & FH      & 88                & \textbf{0.32} & 1.34 \\
    \bottomrule
    \end{tabular}}
    \vspace{-0.6em}
    \caption{{\bf Adding positional embeddings to patch features before pooling them into region tokens gives the best reconstruction and generation.} REN tokens reconstruct poorly because their regions overlap and their features encode little position. Fine-tuning REN makes the number of tokens grow during training ($^\ast$mean token count at the start${\to}$end of training). Pooling patch features within fixed FH regions keeps the token count stable, but the tokens still encode little position. Adding an explicit mask code to each token improves reconstruction but hurts generation, because the DiT must generate these codes precisely.}
    \label{tab:ae-ablation}
\end{table}

\subsection{How to generate without implicit spatial prior?}
\label{sec:generating-layout}
Patch tokens occupy fixed grid positions, given to the DiT through 2D RoPE. Region tokens, in contrast, vary in position, shape, and size from image to image, so the DiT must generate both {\em what} each region contains and {\em where} it lies. We compare multiple ways of generating region tokens, from treating them as an unordered set to anchoring them at random points, as discussed below (Tab.~\ref{tab:layout-problem}).

\boldheader{Generating region tokens as an unordered set}
In the simplest design, the DiT receives no explicit spatial prior and generates region tokens that jointly encode content and position. Since the DiT must denoise both where each region belongs and what it contains, regions can shift across denoising steps, disrupting the generation of consistent content at each location.

\boldheader{Conditioning on a given layout}
We provide the DiT with a layout and train it to generate only the pooled DINOv3 features. Concretely, for each training image, we use its FH segmentation to obtain region centers for 2D RoPE and mask embeddings that are added to the tokens being denoised. This follows the explicit mask encoding design discussed in Sec.~\ref{sec:encoding-layout}. At sampling time, borrowing the layout of a random training image produces remarkably good results, even outperforming RAEv2. However, layouts borrowed from validation images cause a sharp drop in quality, indicating that the DiT memorizes training layouts rather than generalizing to new ones. The approach therefore does not yield a standalone generator, since it requires training-set layouts at sampling time.

\boldheader{Learning to generate layouts}
To remove the dependence on borrowed layouts, we split generation across two DiTs. The first DiT generates layout for the image, and the second generates the region tokens conditioned on the generated layouts. This works, but the two sequential stages nearly eliminates the efficiency advantage of \rengen{} over RAEv2, even though \rengen{} generates $2.9{\times}$ fewer tokens. To avoid the two-stage setup, we also try a single DiT that generates the layout and region tokens jointly, conditioning each step on the layout predicted in the previous step, but this still does not work well.

\boldheader{Random anchor points (ours)}
When pooling patch features with embedded positions (Sec.~\ref{sec:encoding-layout}), we find that although the autoencoder is trained only on FH segments, it reconstructs nearly as well from other partitions: FH segments yield an rFID of 0.320, compared with 0.332 for Voronoi cells and 0.323 for SLIC-style superpixels, both computed from randomly sampled points. We therefore need not predict or borrow an FH layout for generation, and can instead define the region layout from random anchor points. For each training image, we (1) sample $P$ points on the canvas, with $P$ drawn uniformly within $\pm20\%$ of the image's FH segment count; (2) partition the image into their Voronoi cells or SLIC-style superpixels; (3) encode the image under this partition to get the target tokens for the DiT (Fig.~\ref{fig:dit-targets}); and (4) use the anchor positions as 2D RoPE for the DiT. At sampling time, we draw fresh random anchors, with $P$ chosen by the user or drawn from the class's token counts, and the DiT generates region tokens that carry both content and position. This requires no layout model, no training images, and no layout guidance. The DiT thus retains the spatial prior that the unordered set lacks, without having to generate a layout explicitly.

\begin{table}[t]
    \centering
    \small
    \setlength{\tabcolsep}{3pt}
    \resizebox{\linewidth}{!}{
    \begin{tabular}{llcc}
    \toprule
    \makecell[l]{\textbf{Spatial input}\\\textbf{at sampling}} &
    \makecell[l]{\textbf{Training}\\\textbf{regions}} &
    \makecell{\textbf{Sampling}\\\textbf{time/img (ms)}} &
    \textbf{gFID}$\downarrow$ \\
    \midrule
    none (unordered set)                       & FH         & 57.3 & 6.32 \\
    \midrule
    FH layout of a train image (memorized)     & FH         & 57.9 & 1.02 \\
    FH layout of a val image                   & FH         & 58.0 & 3.60 \\
    \midrule
    layout generated by a separate DiT         & FH         & 128.0 & 2.47 \\
    layout generated jointly                   & FH         & 135.9 & 2.29 \\
    \midrule
    random anchor points                       & SLIC       & 57.9 & 1.36 \\
    \rowcolor{RowBlue}
    random anchor points (ours)                & Voronoi    & 57.9 & 1.34 \\
    \bottomrule
    \end{tabular}}
    \vspace{-0.6em}
    \caption{{\bf Randomly fixing region positions with anchor points enables strong generation at low cost.} Without any spatial prior, the DiT trains poorly. Conditioning on real layouts separately leads to memorization and requires real images at sampling time, while generating layouts roughly doubles sampling time and still degrades quality. Random anchor points avoid both and outperform held-out real layouts with a single DiT. At sampling, the DiT receives only the points; the training regions determine how the targets are partitioned around them. Sampling time is measured per image at batch size 64 on one NVIDIA GH200.}
    \label{tab:layout-problem}
    \vspace{-0.6em}
\end{table}

\begin{table}[t]
    \centering
    \small
    \setlength{\tabcolsep}{3pt}
    \resizebox{\linewidth}{!}{
    \begin{tabular}{lccccccc}
    \toprule
    \multirow{2}{*}{\textbf{Method}} &
    \multirow{2}{*}{\makecell{\textbf{Content}\\\textbf{adaptive}}} &
    \multirow{2}{*}{\makecell{\textbf{Controllable}\\\textbf{complexity}}} &
    \multirow{2}{*}{\makecell{\textbf{Frozen}\\\textbf{enc.}}} &
    \multicolumn{2}{c}{\textbf{Reconstruction}} &
    \multicolumn{2}{c}{\textbf{Generation}} \\
    \cmidrule(lr){5-6}\cmidrule(lr){7-8}
    & & & & \textbf{rFID}$\downarrow$ & \textbf{PSNR}$\uparrow$ & \textbf{gFID}$\downarrow$ & \textbf{IS}$\uparrow$ \\
    \midrule
    \multicolumn{8}{c}{\textit{Compact fixed-length tokenizers}} \\
    \midrule
    TiTok~\cite{TiTok}   & \xmark & \xmark & \xmark & 1.71 & 17.73 & 1.97 & 281.8 \\
    MAETok~\cite{chen2025maetok} & \xmark & \xmark & \xmark & 0.48 & \textbf{23.61} & 1.67 & \textbf{311.2} \\
    \midrule
    \multicolumn{8}{c}{\textit{Variable-length tokenizers}} \\
    \midrule
    FlexTok~\cite{FlexTok}       & \xmark & \cmark & \xmark & 1.20 & 15.90 & 1.89 & 275.6 \\
    One-D-Piece~\cite{OneDPiece} & \xmark & \cmark & \xmark & 1.77 & 17.07 & -- & -- \\
    AdaTok~\cite{AdaTok}         & \xmark & \cmark & \xmark & 1.63 & 17.92 & -- & -- \\
    GPSToken~\cite{GPSToken}     & \xmark & \cmark & \xmark & 0.65 & \underline{24.06} & 1.50 & -- \\
    LGM~\cite{LGM}               & \xmark & \cmark & \xmark & 0.59 & 22.80 & 1.86 & \underline{290.7} \\
    \midrule
    \multicolumn{8}{c}{\textit{Content-adaptive tokenization}} \\
    \midrule
    ALIT~\cite{ALIT}     & \cmark & \cmark & \xmark & 3.18 & 18.55 & -- & -- \\
    KARL~\cite{KARL}     & \cmark & \cmark & \xmark & 4.05 & 20.78 & -- & -- \\
    AdaTok~\cite{AdaTok} & \cmark & \xmark & \xmark & 1.50 & 17.78 & -- & -- \\
    BudgetPix$^\S$~\cite{kara2026budgetpix} & \cmark & \cmark & \xmark & -- & -- & 4.75 & 284.0 \\
    \midrule
    \multicolumn{8}{c}{\textit{Representation autoencoders}} \\
    \midrule
    RAEv2~\cite{RAEv2} & \xmark & \xmark & \cmark & \textbf{0.30} & 22.58 & \textbf{1.06} & 255.3 \\
    \rowcolor{RowBlue}
    \rengen{} & \cmark & \cmark & \cmark & \underline{0.32} & 22.23 & \underline{1.34} & 253.7 \\
    \bottomrule
    \end{tabular}}
    \vspace{-0.6em}
    \caption{{\bf\boldmath \rengen{} achieves the best rFID and gFID among methods that vary their token count.} It also reconstructs and generates nearly as well as RAEv2 with $2.6{\times}$ fewer TFLOPs and $2.5{\times}$ less wall-clock time (Fig.~\ref{fig:pareto}). The checkmark columns indicate whether a method's token count varies with image content ({\em content adaptive}), whether a user can choose the token budget of a single model at inference ({\em controllable complexity}), and whether its latents come from a frozen pretrained representation encoder ({\em frozen enc.}). $^\S$Included as the closest prior generator that adapts its compute to image content; it operates in pixel space, so reconstruction metrics do not apply.}
    \label{tab:comparison}
    \vspace{-0.8em}
\end{table}

\subsection{Comparison with latent generative models}
\label{sec:comparison}

We show that \rengen{} reconstructs and generates images on par with leading latent generative models at substantially lower compute and latency. Fig.~\ref{fig:pareto} relates generation quality to compute, and Tab.~\ref{tab:comparison} compares reconstruction and generation in detail against (1) representation autoencoders (RAEs) that use a pretrained DINOv3 encoder, (2) compact tokenizers that encode every image into a fixed number of tokens below the $16{\times}16$ grid, (3) variable-length tokenizers that support several token budgets, and (4) content-adaptive tokenizers whose token count varies per image.

\boldheader{Reconstruction} 
\rengen{} reconstructs nearly as well as RAEv2 (rFID 0.32 vs.\ 0.30 with DINOv3-L) while using $2.9{\times}$ fewer tokens. Among tokenizers that vary their token count, it achieves the lowest rFID, $4.7{\times}$ lower than AdaTok (1.50) and $1.8{\times}$ lower than LGM (0.59). \rengen{}'s PSNR is lower than that of tokenizers trained end-to-end for pixel reconstruction, such as MAETok (22.2 vs.\ 23.6). Qualitatively, however, \rengen{} preserves fine structure as well as RAEv2 with far fewer tokens (Fig.~\ref{fig:reconstruction}).

\boldheader{Generation quality and cost} 
\rengen{} reaches a gFID of 1.34 with 88 tokens on average, 0.28 behind RAEv2 (1.06), and is the strongest model among those that vary their token count. Following RAEv2~\cite{RAEv2}, we sample with internal guidance (scale 2.0 on $t\in[0.6,1]$, 50 Euler steps, no CFG). Fig.~\ref{fig:pareto} places this result against compute. \rengen{} lies below the Pareto frontier of prior methods, reaching its gFID at 9.1 TFLOPs per image, while RAEv2 needs 23.9. The right panel shows the source of this gain. At matched model size, \rengen{} generates $2.9{\times}$ fewer tokens than RAEv2, which yields $2.6{\times}$ fewer TFLOPs and $2.5{\times}$ less wall-clock time per image. These savings come entirely from the shorter token sequence.

\subsection{Efficient scaling to higher resolutions}
\label{sec:scaling-resolution}
Being a region-based model, \rengen{} encodes a $512{\times}512$ image into the same number of tokens as a $256{\times}256$ image, so the DiT cost of generating region latents stays the same across resolutions. Only the decoder's cost grows with resolution, and it only runs once per image rather than at every denoising step. In contrast, for patch-based models like RAEs, doubling the image side quadruples the number of tokens and increases the DiT's attention cost by $16{\times}$. To scale efficiently to higher resolutions, RAE trains a decoder with twice the encoder's patch size ($32$ vs.\ $16$), so that a $256{\times}256$ image is encoded into $256$ tokens and decoded at $512{\times}512$~\cite{RAE}. As Tab.~\ref{tab:scaling-resolution} shows, RAE's approach is both less efficient and less effective than \rengen{}. Moreover, since \rengen{}'s decoder queries are 2D Fourier positional embeddings of the output patch grid, a single model can render images at multiple resolutions (here, $256{\times}256$ and $512{\times}512$) by varying only the number and frequencies of the queries. Patch-based models, in contrast, must retrain (at least the decoder) for each new resolution. We leave a fuller study of multi-resolution decoding and generation to future work.

\begin{table}[t]
    \centering
    \small
    \setlength{\tabcolsep}{4pt}
    \resizebox{\linewidth}{!}{
    \begin{tabular}{lccccc}
    \toprule
    \textbf{Method} &
    \makecell{\textbf{Encoded}\\\textbf{resolution}} &
    \makecell{\textbf{Decoded}\\\textbf{resolution}} &
    \textbf{Tokens} &
    \makecell{\textbf{DiT}\\\textbf{GFLOPs}$\downarrow$} &
    \textbf{rFID}$\downarrow$ \\
    \midrule
    \gr{RAEv2 (for reference)}     & \gr{256} & \gr{256} & \gr{256}            & \gr{135.2}          & \gr{0.30} \\
    RAEv2             & 256 & 512 & 256            & 135.2          & 0.50 \\
    \midrule
    \gr{RenGen (for reference)}    & \gr{256} & \gr{256} & \gr{88}             & \gr{47.5}           & \gr{0.32} \\
    \rowcolor{RowBlue}
    \rengen{}                      & 256 & 512 & \textbf{88}    & \textbf{47.5}  & \underline{0.48}   \\
    \rowcolor{RowBlue}
    \rengen{}                      & 512 & 512 & \textbf{88}    & \textbf{47.5}  & \textbf{0.39} \\
    \bottomrule
    \end{tabular}}
    \vspace{-0.6em}
    \caption{{\bf\boldmath \rengen{} scales to higher resolutions more efficiently than RAEv2.} At $512{\times}512$, \rengen{} reconstructs slightly better than RAEv2 while using about a third of the tokens and DiT compute. Because its token count depends on image content rather than resolution, \rengen{} can also encode the full $512{\times}512$ image into the same 88 tokens, improving reconstruction further at no extra cost. A single \rengen{} model handles both $256{\times}256$ and $512{\times}512$ inputs here.}
    \label{tab:scaling-resolution}
    \vspace{-0.6em}
\end{table}
    
\section{Conclusion}
\label{sec:conclusion}

We introduce \rengen{}, a latent diffusion model built on region tokens. We demonstrate that \rengen{} is content adaptive allowing control over scene complexity. On ImageNet $256{\times}256$, \rengen{} reconstructs images with rFID similar to state-of-the-art representation autoencoders using $2.9\times$ less tokens and generates a gFID of 1.34 with $2.5{\times}$ lower latency. For future work, we plan to explore text-to-image training, more efficient training, and decoding at more resolutions and aspect ratios from a single model. Region-based generation also enables finer control, such as editing the layout of a scene or resizing and rotating individual objects.

\section*{LLM Usage Statement}

We used generative AI tools, such as LLMs, for agentic coding (Claude), paper writing, paper editing, assist with translation, creating figures, and search for relevant related work. We have not used generative AI to develop theoretical model, formulate math claims, proofs, create hypotheses, design experiments or provide feedback on research methodology. We have reviewed all AI-assisted work for correctness.
{
    \small
    \bibliographystyle{ieeenat_fullname}
    \bibliography{main}
}
\clearpage
\appendix
\renewcommand{\thetable}{\Alph{section}\arabic{table}}
\renewcommand{\thefigure}{\Alph{section}\arabic{figure}}

\section*{Appendix}
\cref{sec:editing} shows how \rengen{}'s design gives users direct scene control. Users can add detail to a generated image (\cref{sec:preview}), transplant objects between images (\cref{sec:transplant}), move content within an image (\cref{sec:move}), and inpaint regions (\cref{sec:inpaint}). \cref{app:ablations} presents additional ablations. Finally, we provide implementation details for the autoencoder (\cref{app:ae}) and the point-anchored DiT (\cref{app:dit}), and describe the evaluation and timing protocols (\cref{app:eval}).

\section{Direct scene control with anchor points} 
\label{sec:editing} 
\setcounter{table}{0}\setcounter{figure}{0}
A \rengen{} image is a set of anchor points paired with region tokens, so users can control an image directly by changing this set. Adding anchor points adds detail (\cref{fig:grow}), replacing tokens transplants objects (\cref{fig:swap}), moving points moves content (\cref{fig:shift,fig:shift-gradual}), regenerating tokens from noise inpaints missing regions (\cref{fig:inpainting}), and interpolating tokens blends two objects (\cref{fig:blend}). All controls use the trained model as is, with no extra training, pixel masks, attention manipulation, or inversion. {\bf These controls follow from two properties that patch-based generators lack: (1) token positions are inputs to the generator rather than fixed grid indices, and (2) the number of tokens is not fixed by the architecture.} Notably, generating an image and then editing it with \rengen{} is faster than just generating an image with RAEv2 (\cref{fig:edit-timing}). We demonstrate these controls qualitatively and leave their systematic evaluation, and extensions such as text-guided control, to future work.

\begin{figure}[b]
    \centering
    \includegraphics[width=\linewidth]{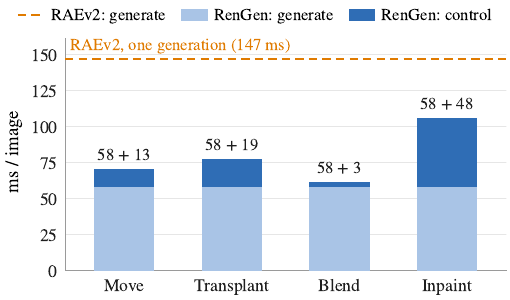}
    \vspace{-1.6em}
    \caption{{\bf Generating an image and then editing it with \rengen{} takes less time than a single RAEv2 generation.} Wall-clock time per image (batch 64, bf16, one GH200). Each bar stacks a \rengen{} generation (light, 50 steps, 88 anchor points) and one control (dark). Moving and transplanting run 10 denoising steps, blending only encodes and decodes, and inpainting regenerates the masked tokens over 50 steps. The dashed line marks one RAEv2 generation (256 tokens, 50 steps, official guidance).}
    \label{fig:edit-timing}
\end{figure}

\subsection{Preview cheaply, then refine}
\label{sec:preview}
\begin{figure}[t]
    \centering
    \includegraphics[width=0.97\linewidth]{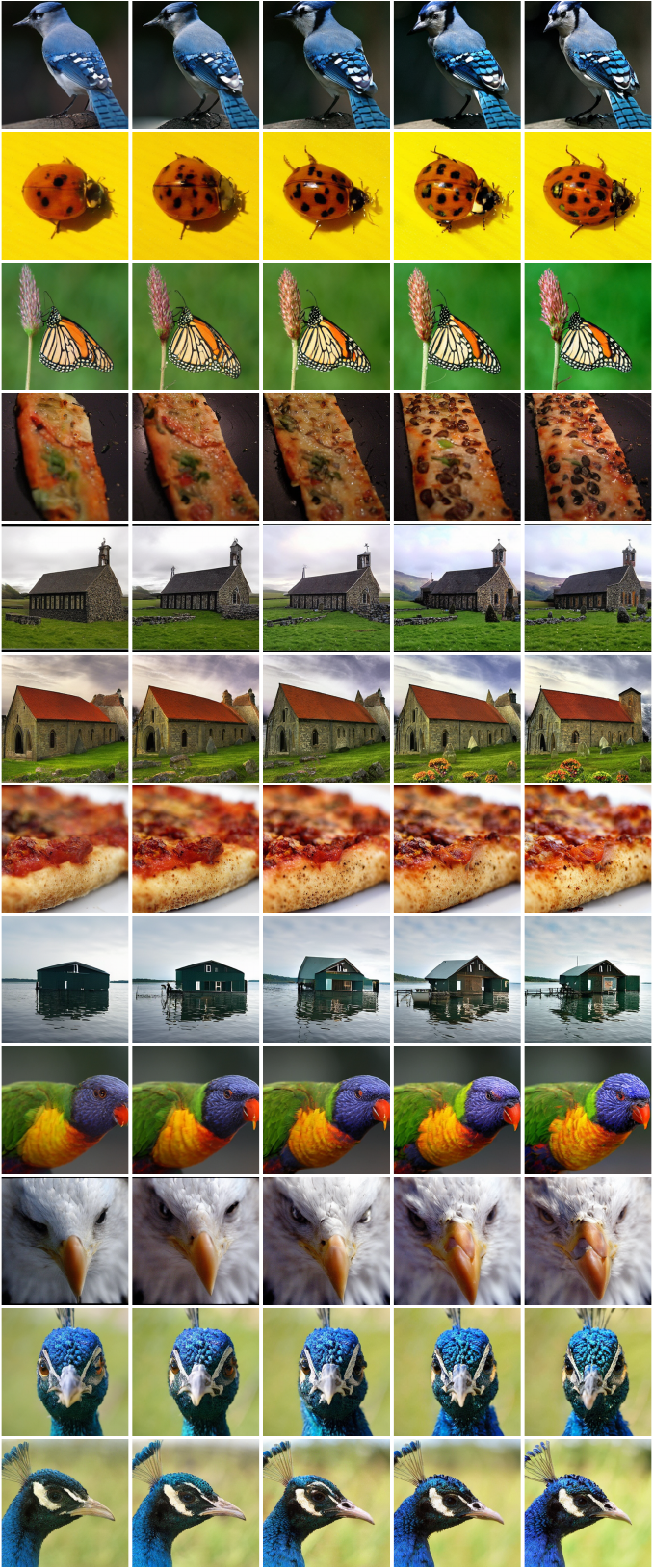}
    \vspace{-0.6em}
    \caption{{\bf Users can preview layouts cheaply with few tokens and then add anchor points to refine the chosen one.} The left column shows images generated with only 32 tokens, cheap enough to sample many candidate layouts. Adding anchor points to a chosen image keeps its layout and adds fine detail.}
    \label{fig:grow}
\end{figure}

Users can sample many cheap previews and pay for full detail only on the one they choose (\cref{fig:grow}). We generate images with only 32 tokens, then add anchor points to a chosen image while keeping its existing points fixed. Each new token starts as a copy of its nearest existing token, and we re-noise all tokens to $t_0\approx0.9$ before finishing sampling with up to 128--148 tokens. The composition, pose, and palette of the preview stay fixed while detail accumulates; e.g., more spots on the ladybug, more feathers on the birds, and more toppings on the pizza. To preview cheaply, patch-based models generate at a lower resolution since their token count is tied to image size, and refining the preview requires a model trained to generate across resolutions. \rengen{} instead previews at full resolution and refines by adding tokens.

\subsection{Transplanting objects between real images} 
\label{sec:transplant}
\begin{figure}[t]
    \centering
    \includegraphics[width=0.96\linewidth]{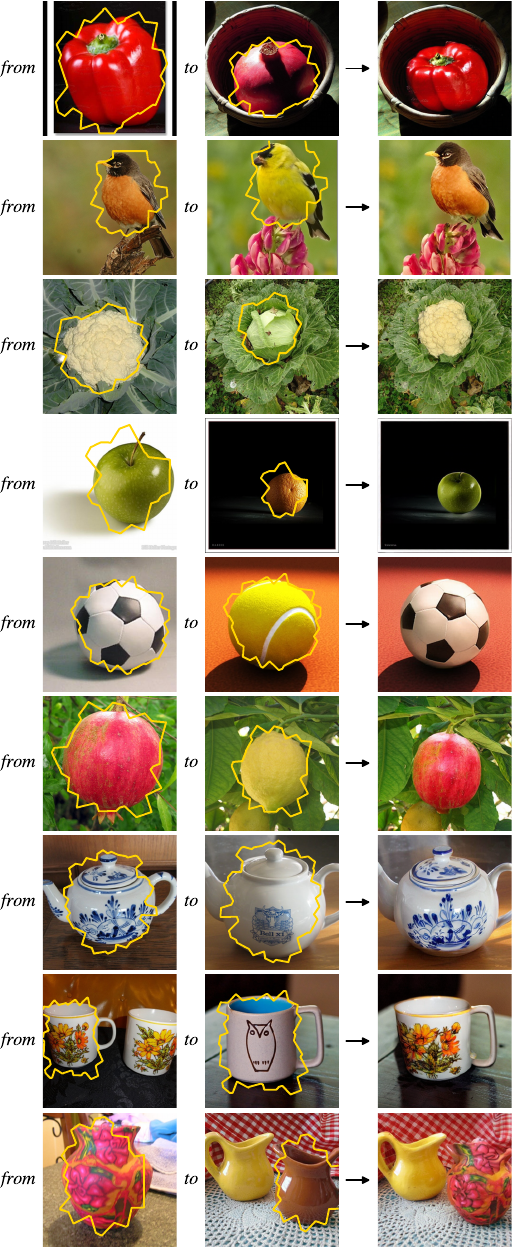}
    \vspace{-0.6em}
    \caption{{\bf Swapping region tokens between images lets users transplant objects, training-free.} The tokens inside the outlined region of the destination are replaced with the source's tokens at those points. The transplanted object adopts the destination's lighting, shadows, and surroundings.}
    \label{fig:swap}
\end{figure}

Users can move an object from one real photo into another (\cref{fig:swap}). We encode both images at the same 128 anchor points, after aligning the source object's bounding box with the destination's. We then replace the destination tokens inside the object with the source tokens, re-noise to $t_0{=}0.67$, and denoise with the source class. The transplanted object adopts the destination's lighting and context. For example, in \cref{fig:swap}, the apple gains a shadow on the dark backdrop, the robin perches on the lupins, and the cauliflower sits among cabbage leaves. The edited region is simply a set of tokens selected by their anchor points, so no pixel mask is needed, and real images need no inversion because the encoder gives their tokens directly. Comparable edits with patch-based diffusion models typically rely on inversion~\cite{mokady2023null} and on pixel masks~\cite{yang2023paint} or attention manipulation~\cite{hertz2023prompt}.

\subsection{Moving content} 
\label{sec:move}
\begin{figure*}[t]
    \centering
    \includegraphics[width=0.89\textwidth]{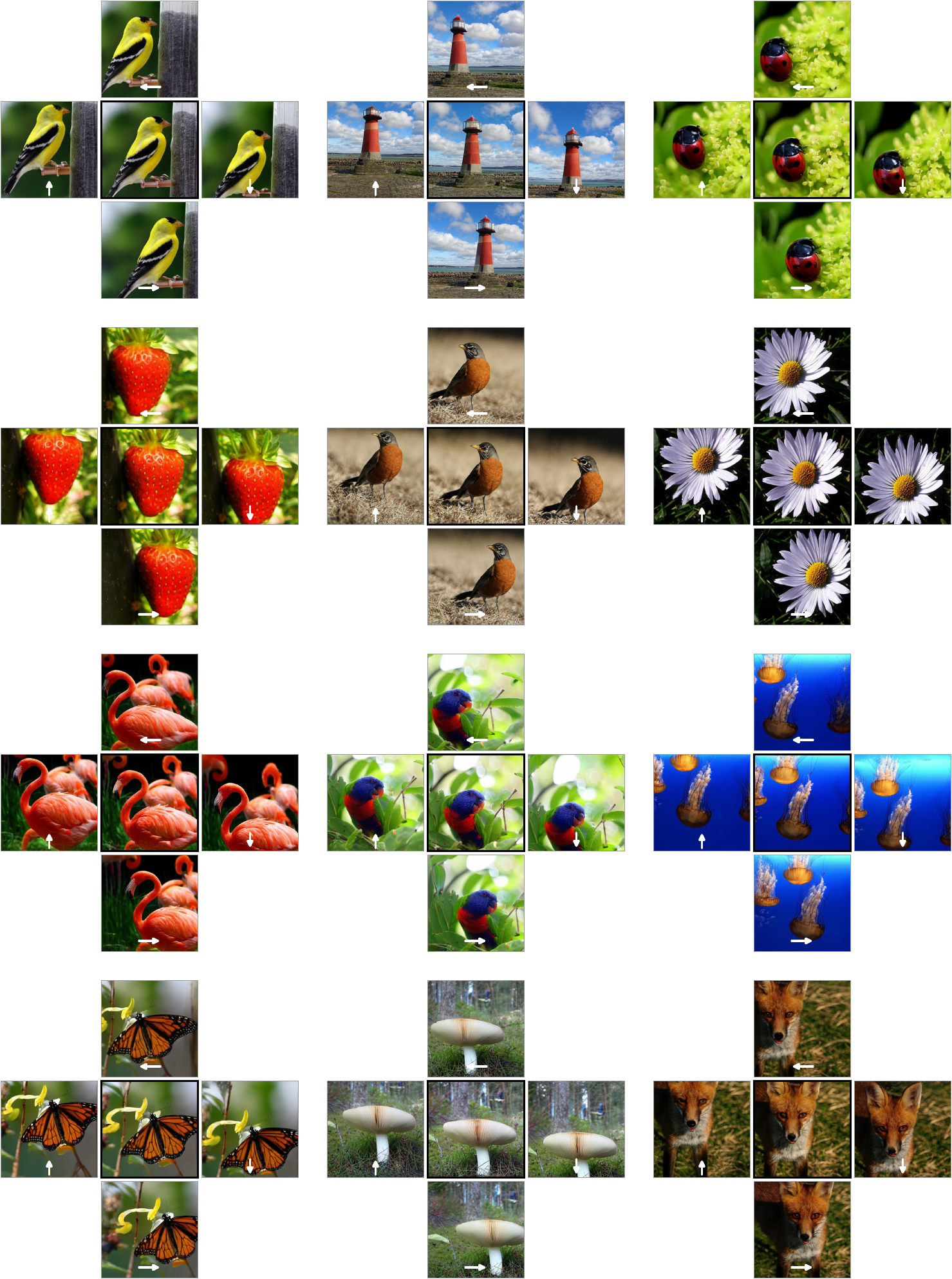}
    \vspace{-0.6em}
    \caption{{\bf Moving the anchor points lets users reposition the scene, training-free.} Each panel's center shows a generated image. Shifting its anchor points along the arrow moves the scene accordingly, and \rengen{} fills in the uncovered border without masks or inpainting.}
    \label{fig:shift}
\end{figure*}

\begin{figure*}[t]
    \centering
    \includegraphics[width=0.87\textwidth]{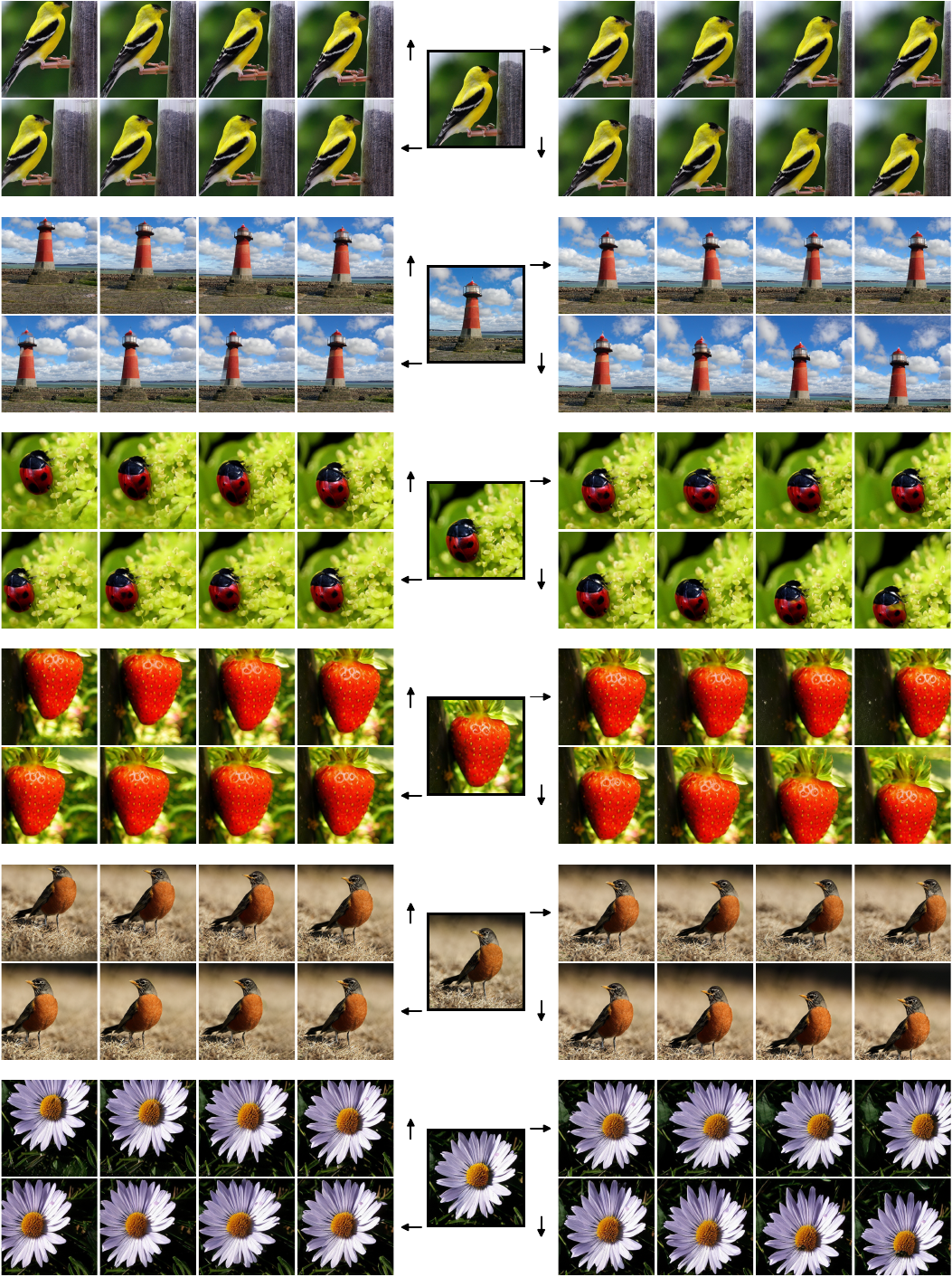}
    \vspace{-0.6em}
    \caption{{\bf Moving the anchor points lets users reposition content continuously, training-free.} Each row shifts the anchor points of the outlined center image by 5 to 20\% of the image width, horizontally (top) or vertically (bottom), using the same noise for every offset. Even a 5\% shift (12.8\,px), smaller than one patch, moves the content smoothly.}
    \label{fig:shift-gradual}
\end{figure*}

Users can move a scene by moving its anchor points. In \cref{fig:shift}, we translate every point by the same offset and re-noise to $t_0{=}0.67$, so only the last 10 of 50 sampling steps run. The object keeps its identity, pose, and colors, and the generator fills in the strip uncovered at the trailing edge. Because anchor points are continuous, the offset need not align with any grid. As shown in \cref{fig:shift-gradual}, steps of 5\% of the image width (12.8 px, less than one 16 px patch) move the content smoothly. In a patch-based model, token positions are fixed by the grid, so content can at best be shifted by whole patches, and the uncovered strip must be filled separately.

\begin{figure*}
    \centering
    \includegraphics[width=\linewidth]{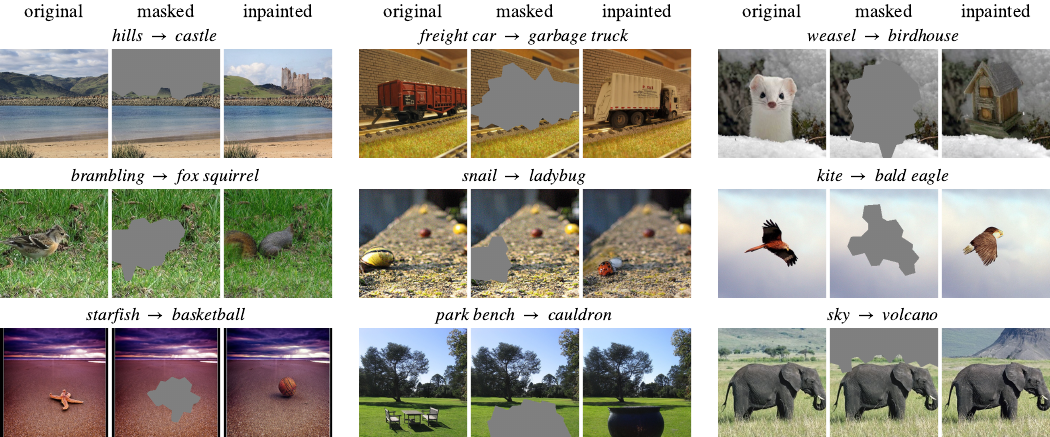}
    \vspace{-1.4em}
    \caption{{\bf Inpainting replaces any region of a real image with an object of a chosen class, training-free.} Tokens in the masked cells (gray) are regenerated with the new class while all others stay fixed. The inserted content matches the scene's scale, lighting, and depth of field.}
    \label{fig:inpainting}
\end{figure*}

\begin{figure*}
    \centering
    \includegraphics[width=\linewidth]{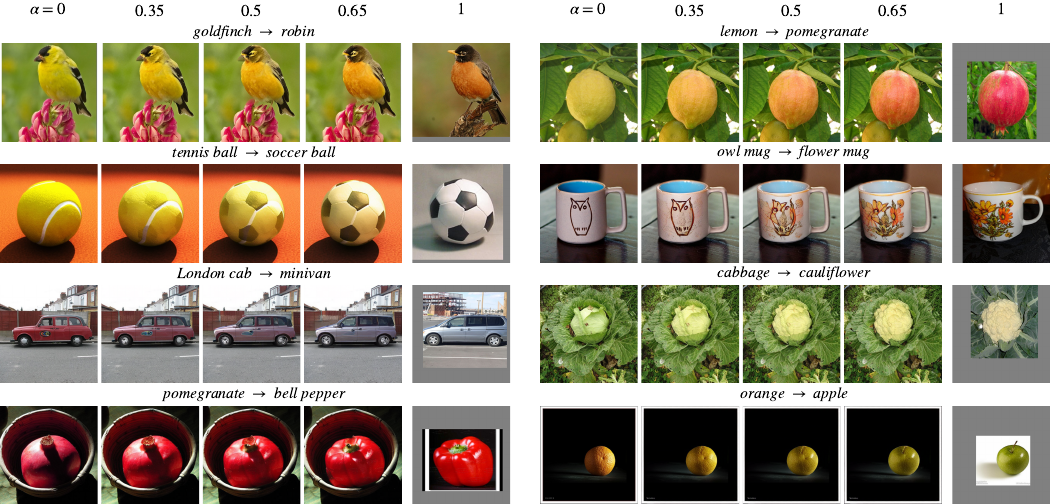}
    \vspace{-1.4em}
    \caption{{\bf Interpolating region tokens blends two objects into plausible hybrids rather than cross-fades.} Tokens inside the destination object ($\alpha{=}0$) are interpolated toward those of the aligned source image ($\alpha{=}1$, right) and decoded directly. Each hybrid keeps the destination's pose, lighting, and background.}
    \label{fig:blend}
\end{figure*}

\subsection{Inpainting}
\label{sec:inpaint}
Users can replace any part of a real image with new content, including an object of a different class, using the same trained generator (\cref{fig:inpainting}). We encode the image at $P$ anchor points and select the cells to replace. The selected tokens start from pure noise and are denoised over the full schedule with a chosen class $c$, while all other tokens are fed in clean at every step and stay unchanged. Through the anchor points, the generator sees both the known content and the exact shape of the hole, and fills the hole to fit its surroundings. In \cref{fig:inpainting}, the inserted objects follow the scene's scale, lighting, and depth of field. Because the hole is a set of tokens rather than a pixel mask, it can take any shape formed by the cells, and no inpainting-specific training or mask conditioning is needed. Patch-based models such as RAEv2 can inpaint in the same way, but they always process the full grid of $256$ tokens.

\subsection{Blending objects}
\label{sec:blend}
Users can blend two objects into a single hybrid by interpolating their tokens (\cref{fig:blend}). We align the source object with the destination object by scaling and translating the source image so that their bounding boxes coincide, and encode both images at the same anchor points, so tokens at the same point describe the same part of the object. Inside the destination object, we interpolate $z_i = (1-\alpha)\, z_i^{\text{dst}} + \alpha\, z_i^{\text{src}}$, keep all other tokens unchanged, and decode directly without sampling. Intermediate values of $\alpha$ produce a single plausible object that keeps the destination's pose, lighting, and shadow, rather than a cross-fade. Most of the change happens for $\alpha$ between about 0.35 and 0.65, suggesting that interpolating region tokens moves between objects along plausible images rather than averaging their appearance.

\section{Additional ablations}
\label{app:ablations}
\setcounter{table}{0}\setcounter{figure}{0}

\subsection{Sensitivity to the region partition}
\label{app:fh-ablation}
\rengen{} uses Felzenszwalb--Huttenlocher (FH) segmentation to define regions for the autoencoder. We test how sensitive reconstruction is to the FH parameters by varying them at inference time, one at a time, with the autoencoder trained at $k{=}100$, $\sigma{=}0.8$, $m{=}32$ (\cref{tab:fh-ablation}). Reconstruction quality changes little across a wide range of settings, even though the number of tokens per image roughly doubles across this range. The scale $k$ has the largest effect on the token count and gives a natural knob for the token budget. Lowering $k$ yields more, finer regions with similar quality, and quality degrades only once regions become coarse ($k{\geq}200$). Smoothing $\sigma$ and the minimum segment size $m$ mainly shift the token count and have a smaller effect on quality. A single autoencoder can therefore operate at different token budgets without retraining, and our default lies in the flat part of this trade-off.

\begin{table}[t]
  \centering
  \small
  \setlength{\tabcolsep}{5pt}
  \begin{tabular}{lcccc}
    \toprule
    \textbf{Setting} & \textbf{Tokens} & \textbf{rFID}$\downarrow$ & \textbf{PSNR}$\uparrow$ & \textbf{SSIM}$\uparrow$ \\
    \midrule
    \multicolumn{5}{c}{\textit{Scale $k$ ($\sigma{=}0.8$, $m{=}32$)}} \\
    \midrule
    $k{=}$25  & 158.4 & 0.328 & 21.44 & 0.604 \\
    $k{=}$50  & 124.0 & \textbf{0.313} & 21.88 & \textbf{0.609} \\
    \rowcolor{RowBlue} $k{=}$100 & 88.3 & 0.321 & \textbf{22.23} & \textbf{0.609} \\
    $k{=}$200 & 56.9  & 0.390 & 21.36 & 0.590 \\
    $k{=}$400 & 33.6  & 0.774 & 19.05 & 0.543 \\
    \midrule
    \multicolumn{5}{c}{\textit{Smoothing $\sigma$ ($k{=}100$, $m{=}32$)}} \\
    \midrule
    $\sigma{=}$0.4 & 107.7 & \textbf{0.312} & 22.03 & 0.608 \\
    \rowcolor{RowBlue} $\sigma{=}$0.8 & 88.3 & 0.321 & \textbf{22.23} & \textbf{0.609} \\
    $\sigma{=}$1.2 & 74.6  & 0.337 & 22.18 & 0.605 \\
    $\sigma{=}$1.6 & 63.4  & 0.366 & 21.84 & 0.598 \\
    \midrule
    \multicolumn{5}{c}{\textit{Minimum size $m$ ($k{=}100$, $\sigma{=}0.8$)}} \\
    \midrule
    $m{=}$8   & 85.7  & 0.324 & 22.15 & 0.607 \\
    $m{=}$16  & 86.7  & 0.323 & 22.18 & 0.608 \\
    \rowcolor{RowBlue} $m{=}$32 & 88.3 & 0.321 & 22.23 & \textbf{0.609} \\
    $m{=}$64  & 88.3  & \textbf{0.320} & \textbf{22.27} & \textbf{0.609} \\
    $m{=}$128 & 71.3  & 0.335 & 22.11 & 0.604 \\
    \bottomrule
  \end{tabular}
  \vspace{-0.6em}
  \caption{{\bf Reconstruction is robust to the FH parameters, so one autoencoder can change its token budget at inference without retraining.} rFID stays within 0.31--0.34 from 74 to 158 tokens per image, and degrades only once regions become coarse ($k{\geq}200$). The autoencoder is trained with $k{=}100$, $\sigma{=}0.8$, $m{=}32$ (shaded), and we vary one parameter at a time at inference. Results on ImageNet-val ($256{\times}256$). Tokens is the mean number of regions per image.}
  \label{tab:fh-ablation}
\end{table}

\subsection{Does generated complexity follow token count?}
\label{app:complexity}
\rengen{} is meant to spend fewer tokens on simple scenes and more on busy ones, so the complexity of a generated image should follow the number of anchor points it receives. We test this by splitting ImageNet-val images into four buckets by FH region count, generating 12.5k images per bucket with the class labels and token counts of that bucket's images, and re-segmenting the generated images with the same FH parameters (\cref{tab:app-complexity}). The mean region count of generated images closely tracks the requested token count, with a rank correlation of 0.94 across all samples. Most generated images fall inside their bucket's region range. The fraction is lower for the middle buckets because their ranges are narrow relative to the per-image spread. In the most complex bucket, generated images are slightly simpler than real ones. Quality stays nearly constant across buckets, so \rengen{} trades scene complexity rather than image quality as the token count changes.

\begin{table}[t]
  \centering
  \small
  \setlength{\tabcolsep}{5pt}
  \resizebox{\linewidth}{!}{
  \begin{tabular}{lccc}
    \toprule
    \textbf{Bucket} & \makecell{\textbf{Real region}\\\textbf{count}} & \makecell{\textbf{Generated region}\\\textbf{count}} & \textbf{In range} \\
    \midrule
    Q1 (1--61)    & 45.6  & 45.9$_{\pm 15.6}$  & 85\% \\
    Q2 (62--84)   & 73.3  & 73.3$_{\pm 14.5}$  & 58\% \\
    Q3 (85--109)  & 96.7  & 94.3$_{\pm 16.5}$  & 54\% \\
    Q4 (110--255) & 133.6 & 122.8$_{\pm 21.4}$ & 72\% \\
    \bottomrule
  \end{tabular}}
  \vspace{-0.6em}
  \caption{{\bf The complexity of generated images matches that of real images.} We report the mean region count of real and generated images (std as subscript) and the fraction of generated images whose region count falls inside the bucket's range (In range).}
  \label{tab:app-complexity}
\end{table}

\section{Autoencoder details}
\label{app:ae}
\setcounter{table}{0}\setcounter{figure}{0}

\cref{tab:app-ae-arch} lists the full autoencoder configuration.

\boldheader{Encoder}
We use a frozen DINOv3 ViT-L/16~\cite{DINOv3}, which maps a $256{\times}256$ image to a $16{\times}16$ grid of patch features. Following RAEv2~\cite{RAEv2}, we aggregate features from seven blocks (every other block, ending with the last). The region encoder projects these features to width $C{=}1152$ and adds random Fourier features $\gamma(\mathbf{u}) = [\sin(2\pi B\mathbf{u}), \cos(2\pi B\mathbf{u})]$ of each patch center $\mathbf{u}$, with $B\in\mathbb{R}^{576\times 2}$ sampled from $\mathcal{N}(0,1)$. Each region query is initialized as the average of the projected features (without positional features) over the region's patches, and is then refined by four cross-attention blocks. The attention logits are biased by $\log m_i$, where $m_i$ is the region's patch-ownership mask, so each token attends only to its own patches.

\boldheader{Decoder}
The decoder uses one query per output patch. Queries are initialized with Mip-NeRF-style integrated positional encodings~\cite{MipNerf} using frequencies $3B$, which suppress frequencies finer than the patch grid, and the same encodings are added before every attention operation. Each of the $L_d{=}14$ layers applies cross-attention to the region tokens, self-attention among the queries, and MLPs. A linear head maps each query to the $16{\times}16{\times}3$ pixels of its patch. We scale the initialization of residual output projections by $1/\sqrt{4L_d}$, without which training stalls and the decoder ignores the tokens.

\boldheader{Training}
We train the region encoder and decoder jointly on ImageNet-1K with Eq.~\ref{eq:ae-loss}, using $\lambda_p{=}1$ and $\lambda_{\text{adv}}{=}0.75$ with VQGAN's adaptive weight~\cite{VQGAN}. The discriminator follows RAE~\cite{RAE}, a frozen DINO ViT-S/8 with trainable heads and DiffAugment~\cite{DiffAugment}. To make the decoder robust to generator errors, we add Gaussian noise to the region tokens during training, with a standard deviation of $\sigma\sim\mathcal{U}(0, 0.8)$ times the per-channel token standard deviation, as in RAE. We train for 48 epochs with AdamW, batch size 512, and bf16 mixed precision. The learning rate follows a cosine schedule from $2{\times}10^{-4}$ to $2{\times}10^{-6}$. Training is completed on a single NVIDIA 4${\times}$GH200 node for all autoencoder experiments.

\begin{table}[t]
    \centering
    \small
    \setlength{\tabcolsep}{4pt}
    \resizebox{\linewidth}{!}{
    \begin{tabular}{ll}
    \toprule
    \multicolumn{2}{c}{\textit{Representation encoder (frozen)}} \\
    \midrule
    Backbone & DINOv3 ViT-L/16 \\
    Feature blocks & 12, 14, \dots, 24 (7 blocks) \\
    Feature grid & $16{\times}16{\times}1024$ \\
    \midrule
    \multicolumn{2}{c}{\textit{Regions}} \\
    \midrule
    Partition & FH, $k{=}100$, $\sigma{=}0.8$, min.\ size 32\,px \\
    Patch ownership & smallest segment with $\geq$25\% coverage \\
    Regions / image & 88.3 mean (5th / 95th pct.\ 34 / 151) \\
    \midrule
    \multicolumn{2}{c}{\textit{Region encoder}} \\
    \midrule
    Width / heads / blocks & 1152 / 16 / 4 \\
    Positional features & 576 random Fourier freqs., $\mathcal{N}(0,1)$ \\
    Parameters & 67.6M \\
    \midrule
    \multicolumn{2}{c}{\textit{Region decoder}} \\
    \midrule
    Width / heads / MLP & 1152 / 16 / 4096 \\
    Layers $L_d$ & 14 \\
    Layer & cross-attn, MLP, self-attn, MLP \\
    Query encoding & integrated PE, frequencies $3B$ \\
    Output & linear, $16{\times}16{\times}3$ per query \\
    Parameters & 415.4M \\
    \midrule
    \multicolumn{2}{c}{\textit{Training}} \\
    \midrule
    Dataset & ImageNet-1K train, $256{\times}256$ \\
    Optimizer & AdamW ($\beta_1, \beta_2 {=}$ 0.9, 0.95) \\
    Learning rate & $2{\times}10^{-4} \to 2{\times}10^{-6}$, cosine \\
    Warmup & 0.25-epoch \\
    Batch size / epochs & 512 / 48 \\
    Loss weights & $\lambda_p{=}1$, $\lambda_{\text{adv}}{=}0.75$ (adaptive) \\
    Discriminator & frozen DINO ViT-S/8, DiffAugment \\
    Token noise & $\sigma \sim \mathcal{U}(0, 0.8)\,\times$ channel std. \\
    EMA decay & 0.9978 \\
    Precision & bf16 \\
    \bottomrule
    \end{tabular}}
    \vspace{-0.6em}
    \caption{\textbf{Autoencoder configuration.}}
    \label{tab:app-ae-arch}
    \vspace{-0.8em}
\end{table}

\section{Point-anchored DiT details}
\label{app:dit}
\setcounter{table}{0}\setcounter{figure}{0}

\cref{tab:app-dit} lists the full DiT configuration.

\boldheader{Architecture}
We use the DiT$^\text{DH}$-XL architecture of RAE~\cite{RAE}, a 28-block transformer followed by a wide, shallow DDT head. Unlike RAE's patch-based model, the input is a variable-length set of $P$ region tokens, padded within a batch and handled with an attention mask. Each token uses its anchor point as a continuous 2D RoPE position, scaled to units of the $16{\times}16$ patch grid, so a point at a patch center receives exactly that patch's rotation angles. The timestep, class label, and token count $P$ are embedded as learnable in-context tokens appended to the sequence. Internal guidance~\cite{IG} reads an auxiliary prediction from an intermediate block through the same head, so it requires no extra network evaluation.

\boldheader{Training targets}
For each training image, we set $P$ to its FH region count, perturbed by a random log-uniform factor and clipped to $[16, 196]$. We sample $P$ points uniformly and refine them with three Lloyd iterations~\cite{Lloyd}. The frozen region encoder then pools the image's DINOv3 features within each Voronoi cell, giving one target token per anchor point (Fig.~\ref{fig:dit-targets} of the main paper). Targets are normalized per channel with training-set statistics, and we group images with similar $P$ into batches to reduce padding.

\boldheader{Training}
We train with flow matching on $x_t = (1{-}t)\,x + t\,\epsilon$, predicting $x$ and computing the loss on the implied velocity over non-padded tokens only. Following RAEv2, timesteps follow a shifted logit-normal distribution. Although \rengen{} trains for more epochs than RAEv2 (120 vs.\ 80), each step costs ${\sim}2.5{\times}$ fewer FLOPs, so its total DiT training compute is roughly 40\% lower. Training is completed on a single NVIDIA 4${\times}$GH200 node for all DiT experiments. 

\boldheader{Sampling}
At sampling time, we draw $P$ from the region count of a random training image of the target class and place the anchor points with the same uniform sampling and Lloyd refinement used in training. No segmentation is run at sampling time. We integrate with a 50-step Euler solver and apply internal guidance over the noisier part of the trajectory.

\begin{table}[t]
    \centering
    \small
    \setlength{\tabcolsep}{4pt}
    \resizebox{\linewidth}{!}{
    \begin{tabular}{ll}
    \toprule
    \multicolumn{2}{c}{\textit{Architecture (DiT$^\text{DH}$-XL)}} \\
    \midrule
    Body width / heads / blocks & 1440 / 20 / 28 \\
    Head width / heads / blocks & 2048 / 16 / 2 \\
    In-context tokens (time / class / $P$) & 4 / 8 / 4 \\
    Positional encoding & 2D RoPE at anchor points \\
    Internal-guidance branch & output of block 8 \\
    Parameters & 876.5M \\
    \midrule
    \multicolumn{2}{c}{\textit{Targets}} \\
    \midrule
    Token count $P$ & FH count $\times\, e^{u}$, $u{\sim}\mathcal{U}(-0.2, 0.2)$ \\
    & clipped to $[16, 196]$ \\
    Anchor points & uniform + 3 Lloyd iterations \\
    Cells & Voronoi, on the $16{\times}16$ patch grid \\
    Normalization & per-channel mean / std \\
    Augmentation & horizontal flip \\
    \midrule
    \multicolumn{2}{c}{\textit{Optimization}} \\
    \midrule
    Objective & flow matching, $x$-pred., $v$-loss \\
    Timesteps & logit-normal(0, 1), shift $s{=}8$ \\
    Label dropout & 0.1 \\
    Optimizer & Muon (2D weights) + AdamW \\
    & $\beta{=}(0.9, 0.95)$, no weight decay \\
    Batch size / grad.\ clip & 1024 / 1.0 \\
    Learning rate & $2{\times}10^{-4}$, linear decay to $2{\times}10^{-5}$ \\
    & over epochs 25--50, then constant \\
    Epochs & 120 \\
    EMA decay & 0.9995 \\
    \midrule
    \multicolumn{2}{c}{\textit{Sampling}} \\
    \midrule
    Solver & Euler, 50 steps, shift $s{=}8.0$ \\
    Guidance & internal, scale 2.1, $t\in[0.6, 1]$ \\
    Anchor points & uniform + 3 Lloyd iterations \\
    \bottomrule
    \end{tabular}}
    \vspace{-0.6em}
    \caption{\textbf{Point-anchored DiT configuration.}}
    \label{tab:app-dit}
    \vspace{-0.8em}
\end{table}

\section{Evaluation protocol}
\label{app:eval}
\setcounter{table}{0}\setcounter{figure}{0}

\boldheader{Generation metrics}
Unless noted otherwise, we compute gFID and IS from 50k samples (50 per class) against ImageNet $256{\times}256$ training-set statistics. Sampling settings are listed in \cref{tab:app-dit}.

\boldheader{Reconstruction metrics}
We compute rFID, PSNR, and SSIM on the 50k ImageNet validation images, center-cropped to $256{\times}256$. rFID is measured against the statistics of the same images.

\boldheader{Complexity buckets (\cref{fig:complexity})}
We sort the validation images by FH region count and split them into four equal buckets of 12.5k images. For each image, we generate one sample with its class label and region count, and compute each bucket's gFID against its 12.5k real images. Because both the sample and reference sets are smaller than in the standard protocol, these values are higher than 50k gFID and are comparable only across buckets. We evaluate BudgetPix on the same buckets with its released JiT-L/16 checkpoint and sampler, and report its token count averaged over the denoising trajectory.

\boldheader{Timing and FLOPs}
We report per-image wall-clock time for end-to-end generation, including anchor placement, 50 guided sampling steps, and decoding. We measure it on one GH200 GPU at batch size 64 in bf16, with $P{=}88$, the mean token count. Baselines are timed with the same setup. We count FLOPs with PyTorch's \texttt{FlopCounterMode} over all DiT evaluations and one decoder pass.

\boldheader{Baselines}
Numbers in \cref{tab:comparison} are taken from the respective papers, except for RAEv2 and BudgetPix, which we evaluate from their released checkpoints. For \cref{fig:reconstruction}, we run the released FlexTok, KARL, RAE, and RAEv2 models with their default reconstruction settings.

\section{Limitations}
\label{sec:limitations}
\rengen{} approaches but does not yet match the gFID of the strongest patch-based models. However, compared to the strongest patch-based model, \rengen{} uses $2.9{\times}$ fewer tokens and $2.5{\times}$ lower latency. Closing this gap without giving up this efficiency remains an open direction. At very low token counts, thin structures and small objects can be lost. This trade-off is inherent to any compressed latent, and in \rengen{} it is easy to avoid, since users can raise the token count for scenes that need it. At sampling time, the token count is either drawn from training images of the class, which keeps generated complexity matched to the data, or set by the user. A model that predicts its own budget from the condition could be more convenient. The controls in \cref{sec:editing} are training-free from the model, with no task-specific training, and leaving room for refinement. Edits follow the cells of their anchor points rather than exact object outlines, transplanting and blending work best for roughly aligned objects of similar shape, and re-noising can slightly alter details near the edit. Finally, we evaluate only class-conditional generation on ImageNet, the standard benchmark for latent generative models. Scaling region tokens to text-to-image generation requires substantially more data and compute, and is a natural next step.

\end{document}